\documentclass{article}
\usepackage{spconf,amsmath,epsfig}
\usepackage{comment,color}
\usepackage{bm}

\title{DDIPNet and DDIPNet+: Discriminant Deep Image Prior Networks for Remote Sensing Image Classification}
\name{Daniel F. S. Santos, Rafael G. Pires, Leandro A. Passos, and Jo\~{a}o P. Papa\thanks{The authors are grateful to S\~ao Paulo Research Foundation (Fapesp) grants \#2013/07375-0, \#2014/12236-1, \#2019/07665-4, and \#2020/12101-0, to the Brazilian National Council for Research and Development (CNPq) \#307066/2017-7 and \#427968/2018-6.}}
\address{S\~{a}o Paulo State University - UNESP\\
	Department of Computing\\
	Bauru, SP - Brazil}
\usepackage{array}
\usepackage{amsfonts}
\usepackage{hyperref}
\begin{document}
%
\maketitle
\begin{abstract}
Research on remote sensing image classification significantly impacts essential human routine tasks such as urban planning and agriculture. Nowadays, the rapid advance in technology and the availability of many high-quality remote sensing images create a demand for reliable automation methods. The current paper proposes two novel deep learning-based architectures for image classification purposes, i.e., the Discriminant Deep Image Prior Network and the Discriminant Deep Image Prior Network+, which combine Deep Image Prior and Triplet Networks learning strategies. Experiments conducted over three well-known public remote sensing image datasets achieved state-of-the-art results, evidencing the effectiveness of using deep image priors for remote sensing image classification.

\end{abstract}

\begin{keywords}
Remote sensing, Deep Image Prior, Triplet Networks, DDIPNet, DDIPNet+
\end{keywords}

\section{Introduction}
\label{s.introduction}

Technological advances toward remote sensing provided a large amount of high-quality imagery data, depicting a detailed overview from the Earth's surface through spatial and spectral resolutions~\cite{xia2017aid}. In a nutshell, such images comprise essential information regarding the land cover, such as rural areas, residential housing, commercial buildings, and vegetation, which is imperative in many real-world problems, e.g., agriculture and city planning.

However, proper segmentation and classification of those regions denote exhausting and cumbersome human tasks, but not for computers. Ulyanov et al.~\cite{ulyanov2018deep} proposed the Deep Image Prior (DIP) approach recently, which employs a generator network to capture low-level image information before any sample-based learning. Besides, Liu et al.~\cite{liu2017scene} proposed the triplet networks, a model that employs weakly labeled images to alleviate the necessity of a massive volume of labeled samples for training.  

This paper proposes two deep learning approaches, i.e., the Deep Image Prior Network (DDIPNet) and the Discriminant Deep Image Prior Network+ (DDIPNet+), which combine DIP modeling and triplet networks strategies with remote sensing image classification. Therefore, the main contributions of this paper are threefold:

\begin{itemize}
    \item to propose two hybrid deep neural network models, i.e., DDIPNet and DDIPNet+;
    \item to provide a novel approach for remote sensing image classification; and
    \item to foster the literature concerning both deep learning and remote sensing image classification.
\end{itemize}

The remainder of this paper is presented as follows. Section~\ref{s.proposal} provides the theoretical background concerning DDIPNet and DDIPNet+. Further, Sections~\ref{s.methodology} and~\ref{s.results} describe the methodology and the experimental results, respectively. Finally, Section~\ref{s.conclusion} states the conclusions and future works.
\section{Proposed Approach}
\label{s.proposal}

\begin{figure}[!htb]
\centering
  \includegraphics[width=3.4in,height=3.4in,keepaspectratio]{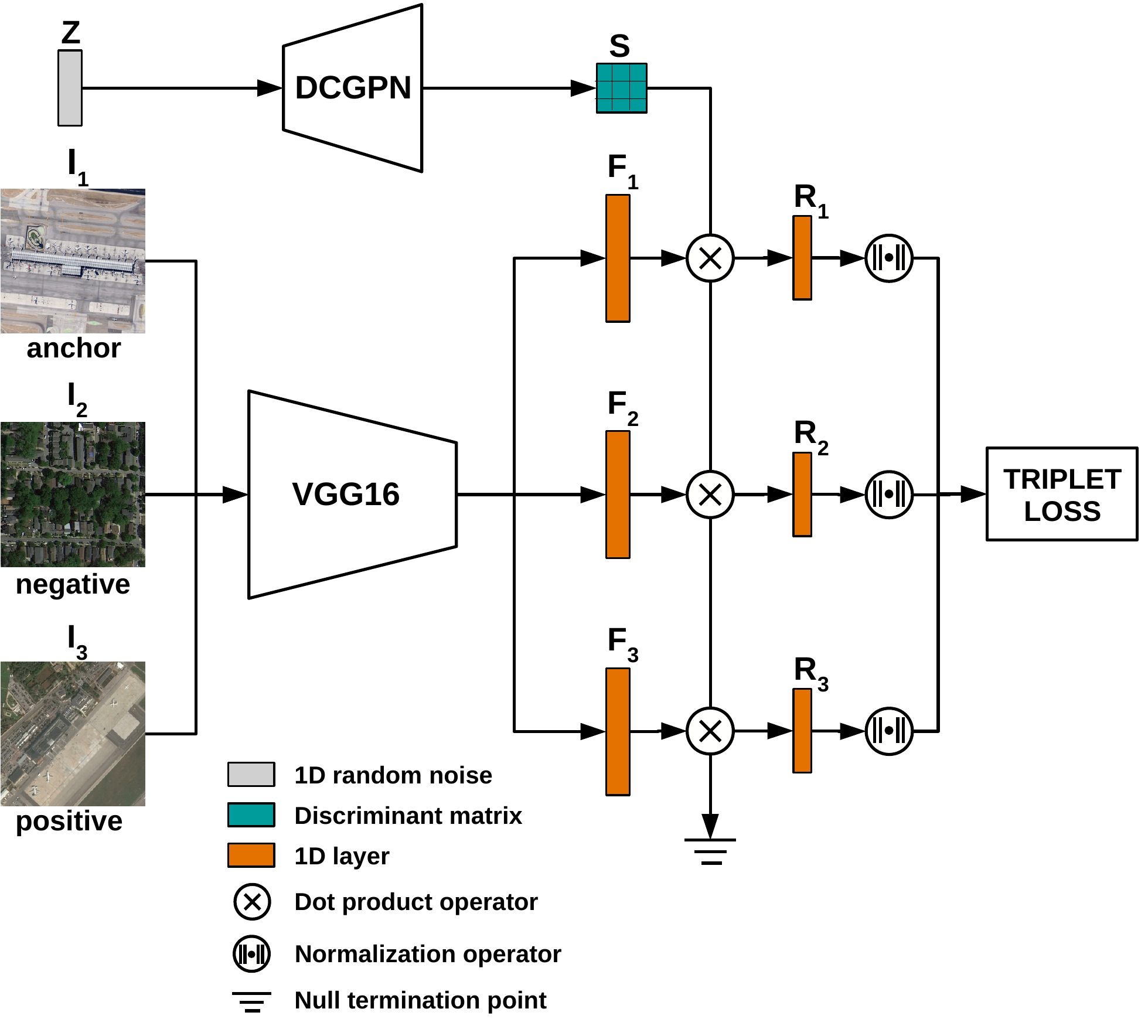}
  \caption{DDIPNet and DDIPNet+ joint optimization architecture configuration.}
  \label{f:ddipnet_arch}
\end{figure}

This section describes DDIPNet and DDIPNet+ architectures, two novel approaches for remote sensing image classification. The models comprise a projective convolutional neural network (VGG16), a Deep Convolutional Generative Prior Network (DCGPN), and a triplet loss function. The difference between DDIPNet and DDIPNet+ stands in the optimization step, in which DDIPNet+ incorporates data augmentation into its triplet network optimization process. Figure~\ref{f:ddipnet_arch} depicts the general idea of the models. 

Given a set of input images ${\cal{I}} = \{\bm{I}_1,\bm{I}_2,\bm{I}_3\}$ such that $\bm{I}_1$ and $\bm{I}_2$ stand for the anchor and negative examples, respectively, and $\bm{I}_3$ denotes the positive sample, the VGG16 neural network is employed to project each of those images into a set of its corresponding low-level feature domain representations ${\cal F} = \{\bm{F}_1,\bm{F}_2,\bm{F}_3\}$. In this work, we considered $\bm{I}_j\in\mathbb{R}^{224\times224}$ and $\bm{F}_j\in\mathbb{R}_{+}^{1\times f}$, where $f = 4,096$ stands for VGG16's last fully connected layer dimension, $\forall j\in\{1,2,3\}$. 

Given a random uniform distributed input $\bm{Z}\in\mathbb{R}^{1\times100}$, the DCGPN network is trained to produce a prior generative model, i.e., a discriminant matrix $\bm{S} \in \mathbb{R}^{f \times c}$, where $c$ stands for the number of classes. The matrix $\bm{S}$ is then used to project ${\cal F}$ into a more compact and separable feature domain $\cal{R}$, such that ${\cal R}=\{\bm{R}_1,\bm{R}_2,\bm{R}_3\}$ and $\bm{R}_j\in \mathbb{R}^{1 \times c}$, $\forall j\in\{1,2,3\}$. Notice that DCGPN follows its original architecture~\cite{radford2015unsupervised}, except for the depth of the last layer, which in this case is equivalent to $c$.

The DDIPNets joint optimization step consists of minimizing a triplet loss function given by: 

\begin{equation}\label{eq:triplet_loss}
L(D_{1}, D_{2};\Theta, \Phi) = max(0, D_{1\{\Theta, \Phi\}} - D_{2\{\Theta, \Phi\}} + m),
\end{equation}
where $\Theta$ and $\Phi$ stand for the VGG16 and DCGPN trainable parameters, respectively, and $m=0.5$ denotes the margin constant.

The term $D_{1\{\Theta, \Phi\}}$ denotes the positive pairwise distance between the anchor feature representation $Q(\bm{R}_{1})$ and the same class feature representation $Q(\bm{R}_{3})$, being computed as follows:

\begin{equation}\label{eq:positive_dist}
D_{1\{\Theta, \Phi\}} = \lVert Q(\bm{R}_{1}) - Q(\bm{R}_{3})\rVert_{2},
\end{equation}
and $D_{2\{\Theta, \Phi\}}$ denotes the negative pairwise distance between the normalized anchor feature representation $Q(\bm{R}_{1})$ and the different class normalized feature representation $Q(\bm{R}_{2})$, being computed as follows:

\begin{equation}\label{eq:negative_dist}
D_{2\{\Theta, \Phi\}} = \lVert Q(\bm{R}_{1}) - Q(\bm{R}_{2})\rVert_{2}.
\end{equation}

Last but not least, $Q(\cdot)$ stands for the non-linear squashing function~\cite{zhang2019remote} and it can be computed as follows:

\begin{equation}
Q(\bm{R}_j) = \frac{\lVert \bm{R}_j \rVert_{2}^{2}}{1 + \lVert \bm{R}_j \rVert_{2}^{2}} \frac{\bm{R}_j}{\lVert \bm{R}_j \rVert_{2}},
\end{equation}
such that $\forall j\in\{1,2,3\}$.

\begin{figure}[!htb]
\centering
  \includegraphics[width=2.8in,height=2.8in,keepaspectratio]{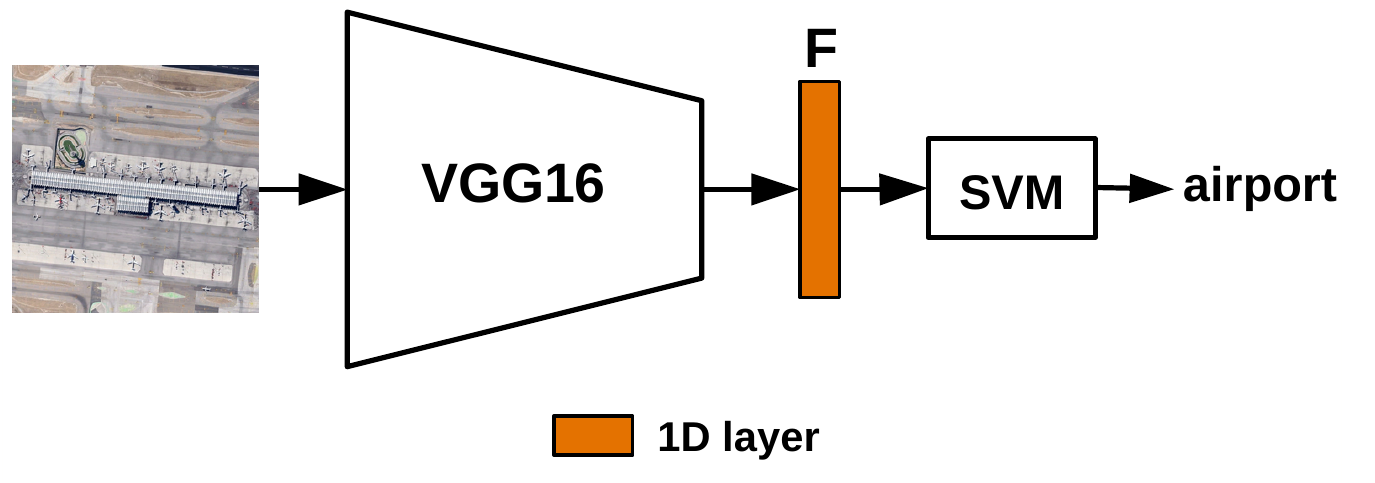}
  \caption{DDIPNet and DDIPNet+ classification architecture.}
  \label{f:ddipnet_svm_arch}
\end{figure}

After VGG16 and DCGPN models have been trained and jointly optimized, the model's output is employed to feed a linear Support Vector Machine (SVM) classifier, as depicted in Figure~\ref{f:ddipnet_svm_arch}.

\section{Methodology}
\label{s.methodology}

We considered three public remote sensing image datasets for a fair evaluation of the proposed approaches, as described bellow:

\begin{itemize}
\item \textbf{UC-Merced}~\cite{yang2010bag}: composed of $21$ classes, each containing $100$ samples of size $256\times256$ pixels, ending up in $2,100$ images.
\item \textbf{AID}~\cite{xia2017aid}: comprises $30$ classes of large-scale aerial scenes totaling $10,000$ images of size $600\times600$ pixels. The dataset is imbalanced.
\item \textbf{NWPU-RESISC45}~\cite{cheng2017remotenwpu}: comprises $31,500$ large-scale aerial scene images of size $256\times256$ equally distributed into $45$ distinct categories.
\end{itemize}

Following the methodology proposed by Zhang et al.~\cite{zhang2019remote}, the models are trained and evaluated over two distinct splitting scenarios. Moreover, different ratios are considered for each dataset, i.e., UC-Merced is trained first with $80\%$ of the samples and further $50\%$ of the instances. Regarding AID dataset, the training folds are built using $50\%$ and $20\%$ of the samples, and finally, NWPU-RESISC45 considers $20\%$ and $10\%$ of the samples for training. For statistical purposes, the procedure is performed during $10$ executions, and the average accuracies are computed for each model.

The training steps are performed as follows:

\begin{enumerate}
\item Both VGG16 (previously trained over ImageNet) and DCGPN networks are jointly trained during a maximum period of $50$ epochs to minimize Equation~\eqref{eq:triplet_loss}. The training procedure considers batches of size $32$, Adam optimizer, and learning rates of $10^{-6}$ and $10^{-4}$ assuming VGG16 and DCGPN models, respectively.
\item Further, the VGG16 network projects the training images into a feature domain through its last fully connected layer.
\item A linear Support Vector Machines (SVM) classifier is trained using the projected features as input\footnote{\url{https://www.csie.ntu.edu.tw/~cjlin/liblinear}}. Notice the default parameters are used according to Xia et al.~\cite{xia2017aid}.
\end{enumerate}
The evaluation process is performed by submitting the test images through the models and measuring the average accuracies obtained in the process.

\section{Experimental Results}
\label{s.results}

Table~\ref{t.overall_accuracy_ucmerced} presents the results over UC-Merced dataset concerning the proposed models against some state-of-the-art approaches. Considering the $80\%$ training ratio, DDIPNet overcomes four-out-of-twelve techniques and presented comparable results concerning another three ones. Over the $50\%$ training ratio, it overcame VGG-16, TEX-Net-LF, and VGG-16-CapsNet, in a total of three-out-of-six baselines. Concerning DDIPNet+, one can observe even better results, outperforming eight-out-of-twelve and four-out-of-six baselines considering $80\%$ and $50\%$ training ratios, respectively.

\begin{table}[hbt!]
\renewcommand{\arraystretch}{1.1}
\centering
\caption{Overall accuracies ($\%$), standart deviations, and rank positions in parenthesis considering UC-Merced dataset.}
\scalebox{.75}{
\begin{tabular}{ccc}
\hline
\textbf{Method}  & \textbf{$\textbf{80\%}$ Training Ratio} & \textbf{$\textbf{50\%}$ Training Ratio} \\ \hline
VGG-16~\cite{xia2017aid}  & $95.21 \pm 1.20$ $(14)$  & $94.14 \pm 0.69$  $(8)$\\
TEX-Net-LF~\cite{anwer2018binary}  & $96.62 \pm  0.49$ $(13)$  & $95.89 \pm 0.37$  $(6)$\\
LGFBOVW~\cite{zhu2016bag}  & $96.88 \pm 1.32$ $(12)$ & /  \\
Fine-tuned GoogLeNet~\cite{castelluccio2015land}  & $97.10$ $(11)$ & /  \\
Fusion by addition~\cite{chaib2017deep}  & $97.42 \pm 1.79$ $(9)$ & /  \\
CCP-net~\cite{qi2018concentric}  & $97.52 \pm 0.97$ $(8)$ & /  \\
Two-Stream Fusion~\cite{yu2018two}  & $98.02 \pm 1.03$ $(7)$ & $96.97 \pm 0.75$  $(4)$\\
DSFATN~\cite{gong2018deep}  & $98.25$ $(6)$ & /  \\
Deep CNN Transfer~\cite{hu2015transferring}  & $98.49$ $(4)$ & /  \\
GCFs+LOFs~\cite{zeng2018improving}  & $99.00 \pm 0.35$ $(2)$ & $97.37 \pm 0.44$  $(2)$\\
VGG-16-CapsNet~\cite{zhang2019remote}  & $98.81 \pm 0.22$ $(3)$ & $95.33 \pm 0.18$ $(7)$  \\
Inception-v3-CapsNet~\cite{zhang2019remote}  & $\textbf{99.05} \boldsymbol\pm \textbf{0.22}$ $\textbf{(1)}$  & $\textbf{97.59} \boldsymbol\pm \textbf{0.16}$ $\textbf{(1)}$ \\
DDIPNet (ours) & \hspace{.1cm} $97.28 \pm 0.75$ $(10)$ & $96.00 \pm 0.70$ $(5)$\\ 
DDIPNet+ (ours) & $98.28 \pm 0.64$ $(5)$ & $97.28 \pm 0.58$ $(3)$ \\ \hline
\end{tabular}}
\label{t.overall_accuracy_ucmerced}
\end{table}

Table~\ref{t.overall_accuracy_aid} presents the results over AID dataset. Considering the $50\%$ training ratio, DDIPNet outperformed three-out-of-seven techniques, while DDIPNet+ also obtained better results surpassing five-out-of-seven of them. Similar results were obtained over the $20\%$ training ratio scenario.

\begin{table}[hbt!]
\renewcommand{\arraystretch}{1.1}
\centering
\caption{Overall accuracies ($\%$), standart deviations, and rank positions in parenthesis considering AID dataset.}
\scalebox{.75}{
\begin{tabular}{ccc}
\hline
\textbf{Method}  & \textbf{$\textbf{50\%}$ Training Ratio} & \textbf{$\textbf{20\%}$ Training Ratio} \\ \hline
VGG-16~\cite{xia2017aid}  & $89.64 \pm 0.36$  $(9)$  & $86.59 \pm 0.29$  $(8)$\\
TEX-Net-LF~\cite{anwer2018binary}  & $92.96 \pm  0.18$ $(7)$ & $90.87 \pm 0.11$  $(6)$\\
Fusion by addition~\cite{chaib2017deep}  & $91.87 \pm 0.36$ $(8)$ & /  \\
Two-Stream Fusion~\cite{yu2018two}  & $94.58 \pm 0.25$ $(5)$ & $92.32 \pm 0.41$  $(4)$\\
GCFs+LOFs~\cite{zeng2018improving}  & $\textbf{96.85} \boldsymbol\pm \textbf{0.23}$ $\textbf{(1)}$ & $92.48 \pm 0.38$  $(3)$\\
VGG-16-CapsNet~\cite{zhang2019remote}  & $94.74 \pm 0.17$ $(4)$ & $91.63 \pm 0.19$  $(5)$\\
Inception-v3-CapsNet~\cite{zhang2019remote}  & $96.32 \pm 0.12$ $(2)$ & $\textbf{93.79} \boldsymbol\pm \textbf{0.13}$ $\textbf{(1)}$  \\
DDIPNet (ours) & $93.91 \pm 0.47$ $(6)$ & $89.88 \pm 0.49$ $(7)$\\ 
DDIPNet+ (ours) & $95.31 \pm 0.22$ $(3)$ & $92.59 \pm 0.52$ $(2)$ \\ \hline
\end{tabular}}
\label{t.overall_accuracy_aid}
\end{table}

Considering NWPU-RESISC45 dataset (Table~\ref{t.overall_accuracy_nwpu}), one can observe that DDIPNet+ was capable of overcoming Fine-tuned VGG-16 and VGG-16-CapsNet considering same training ratios.

\begin{table}[hbt!]
\renewcommand{\arraystretch}{1.1}
\centering
\caption{Overall accuracies ($\%$), standart deviations, and rank positions in parenthesis considering NWPU-RESISC45 dataset.}
\scalebox{.75}{
\begin{tabular}{ccc}
\hline
\textbf{Method}  & \textbf{$\textbf{20\%}$ Training Ratio} & \textbf{$\textbf{10\%}$ Training Ratio} \\ \hline
Fine-tuned VGG-16~\cite{xia2017aid} & $90.36 \pm 0.18$ $(4)$ & $87.15 \pm 0.45$ $(3)$ \\
Triple networks~\cite{liu2017scene}   & $92.33 \pm 0.20$ $(2)$ & /  \\
VGG-16-CapsNet~\cite{zhang2019remote} & $89.18 \pm 0.14$ $(5)$ & $85.08 \pm 0.13$  $(4)$\\
Inception-v3-CapsNet~\cite{zhang2019remote} & $\textbf{92.60} \boldsymbol\pm \textbf{0.11}$ $\textbf{(1)}$ & $\textbf{89.03} \boldsymbol\pm \textbf{0.21}$ $\textbf{(1)}$ \\
DDIPNet (ours) & $88.33 \pm 0.43$ $(6)$ & $83.67 \pm 0.64$ $(5)$\\ 
DDIPNet+ (ours) & $91.54 \pm 0.31$ $(3)$ & $88.14 \pm 0.42$ $(2)$ \\ \hline
\end{tabular}}
\label{t.overall_accuracy_nwpu}
\end{table}

\subsection{Discussion}
\label{ss.discussion}

Summarizing Tables~\ref{t.overall_accuracy_ucmerced}~to~\ref{t.overall_accuracy_nwpu}, we can highlight that DDIPNet showed better results than pre-trained VGG-16, and also than more complex texture-based and visual word-based techniques, such as TEX-Net-LF and LGFBOVW. Meanwhile, DDIPNet+ surpassed the state-of-the-art CapsNet concerning the same VGG-16 backbone. Besides, DDIPNet+ figures a primary advantage over CapsNet since its DCGPN module and discriminant matrix are not incorporated into the final classifier model, thus being quite faster for prediction.

\subsection{Ablation Study}
\label{ss.ablation}

\begin{figure}[!htb]
\centering
  \includegraphics[width=2.9in,height=2.9in,keepaspectratio]{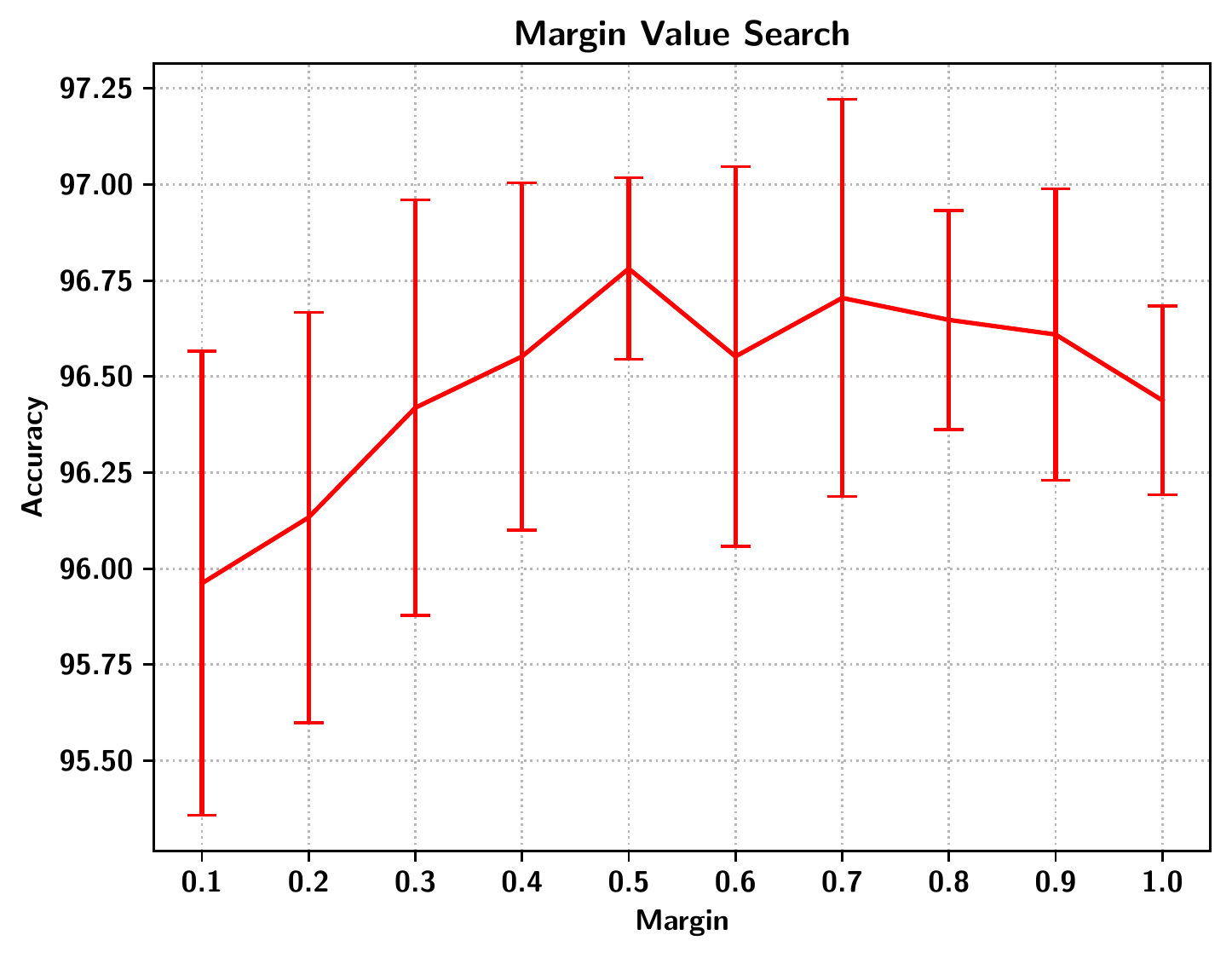}
  \caption{Margin grid search process conducted over UC-Merced~\cite{yang2010bag} dataset, regarding $50\%$ training samples. We considered margin variations of $0.1$, and the searches were executed by five rounds of $15$ training epochs each.}
  \label{f:ddipnet_margin_search}
\end{figure}

As mentioned earlier in Section~\ref{s.proposal}, the proposed approaches figure the main hyperparameter $m$, i.e., the margin constant. We conducted an ablation study to find out the best value. For the sake of space, we present an ablation study over UC-Merced dataset only. Figure~\ref{f:ddipnet_margin_search} depicts such a study, in which $m=0.5$ showed a suitable choice concerning the trade-off between accuracy and robustness (standard deviation).

\section{Conclusion}
\label{s.conclusion}

This work proposes two image classification approaches named DDIPNet and DDIPNet+, which combine Deep Image Prior and triplet network optimization strategies to cope with remote sensing image classification. The proposed approaches showed promising results in three public datasets, overcoming state-of-the-art techniques in some situations. Besides, DDIPNet and DDIPNet+ figure a lighter training and classification steps than their counterparts. Regarding future works, we intend to improve DDIPNet and DDIPNet+ thourgh modifications in the VGG-16 backbone. 

\bibliographystyle{IEEEbib}
\bibliography{ms}

\end{document}